# Autonomous Reality Modelling for Cultural Heritage Sites employing cooperative quadrupedal robots and unmanned aerial vehicles

Nikolaos Giakoumidis[1,2], Christos-Nikolaos Anagnostopoulos [2,*]

[1] KINESIS Lab, Core Technology Platforms, New York University Abu Dhabi, United Arab Emirates
  Affiliation 1; giakoumidis@nyu.edu
[2] Intelligent Systems Lab, Cultural Technology and Communication, University of the Aegean, Greece
  Affiliation 2; canag@aegean.gr

Correspondence: canag@aegean.gr; Tel.: (+30 2251036624)

**Abstract:** Nowadays, the use of advanced sensors, such as terrestrial 3D laser scanners, mobile LiDARs and Unmanned Aerial Vehicles (UAV) photogrammetric imaging, has become the prevalent practice for 3D Reality Modeling and digitization of large-scale monuments of Cultural Heritage (CH). In practice, this process is heavily related to the expertise of the surveying team, handling the laborious planning and time-consuming execution of the 3D mapping process that is tailored to the specific requirements and constraints of each site. To minimize human intervention, this paper introduces a novel methodology for autonomous 3D Reality Modeling for CH monuments by employing autonomous biomimetic quadrupedal robotic agents and UAVs equipped with the appropriate sensors. These autonomous robotic agents carry out the 3D RM process in a systematic and repeatable approach. The outcomes of this automated process may find applications in digital twin platforms, facilitating secure monitoring and management of cultural heritage sites and spaces, in both indoor and outdoor environments.

## 1. Introduction

### 1.1. Problem statement

During recent years, Reality Modeling (RM) technologies including cutting-edge sensor technologies like Terrestrial and Aerial (using drones) Laser Scanning, have found a wide and prominent purpose in the field of Cultural Heritage (CH) modeling, recording and management. However, RM of CH is a constant challenge for surveyors, since it is a manual-driven, laborious and time-consuming process. The scanning path and sensor's positioning are mostly depended on the surveyor's experience, intuition and perception, since an automatic and systematic procedure does not exist. Taking into consideration the natural environment that surrounds CH sites, the challenges become even more complex. Specifically, for the acquisition of a complete 3D Reality model of a large-scale cultural space, multiple manual terrestrial laser scans (TLS) and aerial scans with UAVs (drones) must be performed. In this manual procedure, the scanning path/strategy and the identification of the scanner position, or Next Best View task as presented in the literature, is mostly depended on the operator's experience and perception. As a result, an optimization of the NBV problem to capture efficiently a large-scale complex sites or monuments in dynamic environments (e.g., due to growing or changing vegetation) is quite important to minimize the surveying time and scanning cost. Although the NBV problem is crucial, efficiency and optimality have not been considered qualitatively and explicitly so far in the literature, and thus, in some cases, the surveying process takes usually longer than necessary, since some regions are overlapped unnecessary and extra positionings are planned just to be on the safe side.

### 1.2. Proposed solution - objectives

To improve the above laborious task, a technological platform for autonomous Reality Modeling is proposed in this communication paper. The aim is to create a complete, autonomous, systematic and optimized 3D scanning strategy, to minimize cost and speed up the overall 3D RM process. To achieve the above, two (2) scientific pillars are crucial to be implemented: a) efficient training for the navigation and operation of autonomous robotic agents carrying RM sensors and b) identify the optimal configuration for Terrestrial and aerial survey positions and path, which should maximize scanning area coverage, minimize the number of scanning positions (NBV problem) and identify an optimum scanning/modeling path.

### 2. Autonomous Robotic Agents and NBV: A brief overview

Autonomous 3D scanning using quadrupedal robots has gained significant attention in recent years. These robots have demonstrated the capability to perform agile and dynamic motions, adapt and navigate efficiently to their environment, making them valuable tools for various autonomous applications. Quadrupedal robots have been utilized for 3D scanning strategies to generate a complete set of point clouds of physical objects through multi-view scanning and data registration [1-3]. Additionally, the development of teleoperated quadrupedal robots enhanced with manipulator arms and grippers has expanded the capabilities of these robots for various applications, including 3D scanning [4].

Furthermore, the control of quadrupedal robots has seen experimental success in achieving robust and agile locomotion, with research focusing on model predictive control for dynamic motions in 3D space [5]. The utilization of representation-free model predictive control and exact feedback linearization has been implemented on quadrupedal robots, contributing to the stabilization of periodic gaits for quadrupedal locomotion [6]. Additionally, the application of hybrid dynamical systems has achieved physically effective and robust instances of all virtual bipedal gaits on quadrupedal robots [7].

On the other hand, the NBV problem is a multidisciplinary research area that encompasses computer science, robotics, artificial intelligence, and machine learning, with the goal of enabling autonomous systems to make informed decisions about viewpoint selection for 3D data acquisition and reconstruction. Next Best View (NBV) estimation in 3D environments is a critical aspect of autonomous data acquisition and 3D reconstruction. It involves determining the most informative viewpoint for a sensor or robotic system to capture data that maximizes the information gain while considering factors such as occlusions, completeness, and reconstruction quality. Researchers have proposed various approaches for NBV estimation, including probabilistic frameworks [8], volumetric information gain metrics [9-10], guided NBV for 3D reconstruction of large complex structures using Unmanned Aerial Vehicles (UAVs) [11] and strategies for selecting the next best view based on ray tracing and already available BIM information [12].

Furthermore, the NBV problem has been addressed in the context of surface reconstruction of large-scale 3D environments with multiple UAVs [13], effective exploration for Micro Aerial Vehicles (MAVs) based on expected information gain [14]. These approaches leverage techniques such as reinforcement learning [15], feature tracking, and reconstruction for NBV planning [16], and history-aware autonomous 3D exploration [17]. They aim to address the challenge of selecting the most informative viewpoint for 3D mesh refinement [18].

## 3. Autonomous Reality Modeling for Cultural Heritage

### 3.1 Equipment

The proposed ARM4CH platform consists of a suite of specialized equipment aiming at fully automated and cooperative robotic procedures in the field of 3D Reality Modelling in Cultural Heritage field. Specifically, it consists of quadrupedal robotic agents (i.e. Spot by Boston Dynamics) that may navigate terrain with excellent levels of mobility, performing automated operations, tasks and data capture in a safe, accurate and frequently manner. Each unit is accompanied with Software Development Kit (SDK) to create custom controls, program autonomous missions, and integrate sensor inputs into data analysis tools. It can carry multiple sensors such as 3D scanners, LiDAR and RGB/Depth/Thermal cameras. Besides those sensors it possesses a variety of capabilities to perceive, understand, and navigate its surroundings using Processing Units with Inertial Measurement Unit (IMUs), Global Positioning System (GPS), Simultaneous Localization and Mapping (SLAM) and Odometry technologies. Integrating all these components, a robot is enabled to navigate autonomously in unknown environments, taking into account the real-time data it gathers from its sensors while making intelligent decisions based on its objectives.

The platform is completed with Autonomous flying Unmanned Aerial Vehicles (UAV) that are used in cases that ground scanning is not possible. Aerial surveys are performed using drones that carry high spatial resolution LiDAR and camera in autonomous operation mode with advanced obstacle avoidance for easy reality capture from above.

The ground robotic agents may be programmed to operate in co-operative mode with the aerial unmanned vehicles (drones). Ground robots differ from drones in important ways. They capture data from other perspectives – for example, they can enter buildings or confined spaces and capture close-up images or videos at ground level. And because ground robots are not constrained by airspace flying regulations, they can be utilized in areas where drones are not permitted.

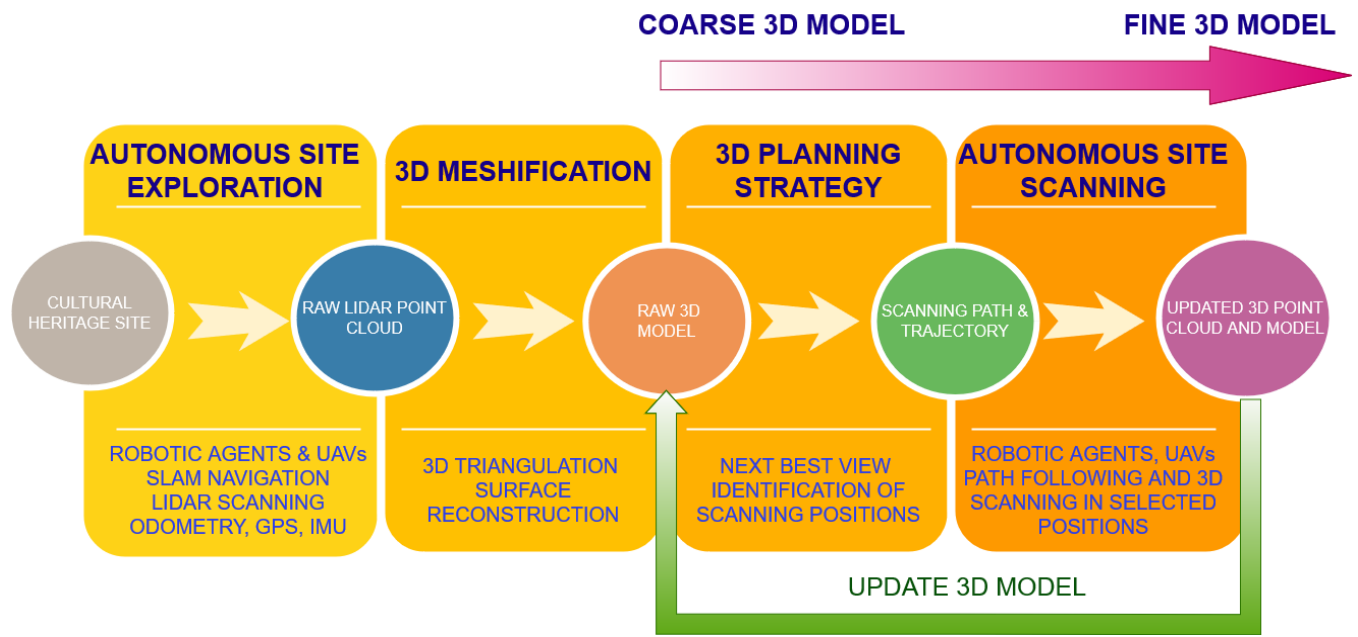

*Figure 1. The four stages of the overall ARM4CH methodology.*

### 3.2 Methodology core

The methodology proposed in this communication paper comprises of four stages/steps.

1. *Acquisition of a coarse 3D LiDAR map*

In this step, a first mapping of the heritage site is carried in order to acquire the basic geometry of the place. Robotic agents (e.g. quadrupedal robot or UAVs) equipped with Simultaneous Localization and Mapping (SLAM) technology and LiDAR sensor dynamically estimate a basic 3D map of the surveyed environment, either indoors (quadrupedals) or outdoors (quadrupedals and UAVs), in the form of a global point cloud. This dynamic operation comes with the limitations of low point cloud resolution, high noise due to motion distortion and inability to record RGB-mapped point cloud. However, as a great advantage, LiDAR sensors calculates rapidly a coarse 3D topological map of the surveyed area, providing a solid ground for the execution of the next steps.

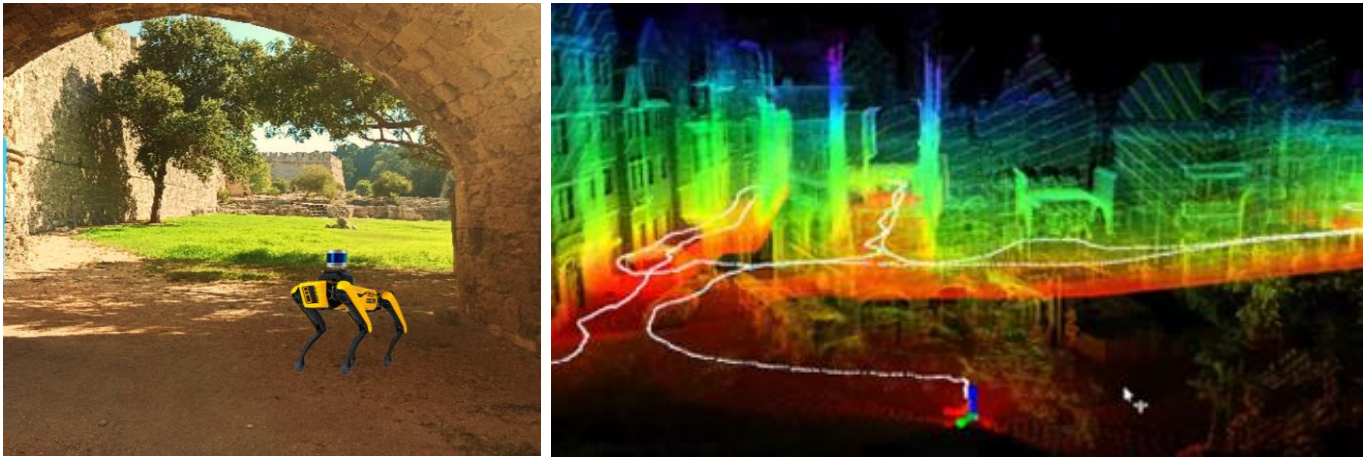

*Figure 2. Autonomous navigation of robotic agents and LiDAR point cloud acquisition.*

## 2. 3D meshification of the global point cloud

3D meshification from point clouds involves various techniques such as parameterization, surface reconstruction, and the utilization of structural or shape priors. These approaches are essential for addressing the challenges posed by the irregular and unstructured nature of point cloud data, ultimately leading to the generation of accurate and detailed polygon meshes. Basic algorithms include the well-known Poisson Surface Reconstruction [19] to generate mesh models using unstructured point clouds [20] and recent learning-based techniques such as Deep Marching Cubes [21] based on Marching Cubes [22]. The overall 3D mesh, which still may suffer from local sparseness, distortion, weak texture and low accuracy, will be the initial model to calculate the optimum number and scanning position for the overall site survey.

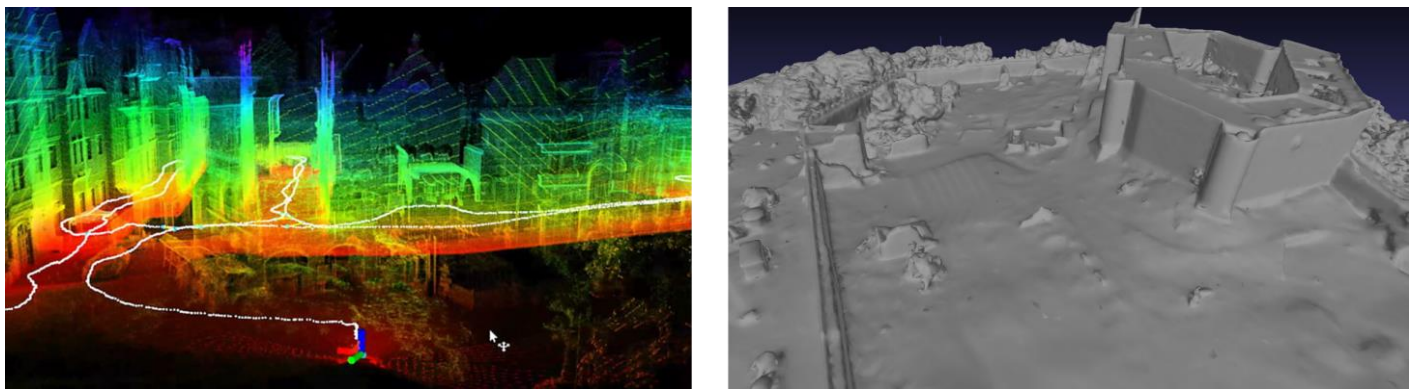

*Figure 3. 3D meshification of the coarse point cloud.*

## 3. Maximization of scanning coverage on the prior 3D mesh

This step is a core algorithmic procedure that aims at solving the NBV problem and defining the scanning positions that integrates successfully the reality modeling planning and minimize the blind/occluded regions. It is a model-based approach running in software (virtual environment) on the bases of the prior model obtained from the coarse 3D LiDAR scan of the site, to define a planning strategy for the identification of the proper scanning positions.

To this context, various methodologies are used that describe the uncertainty of an observation point (candidate scanning position) and finally propose the optimum locations at which the acquired 3D surveying information is expected to be maximized (e.g. using entropy models of LiDAR point cloud volumes) [23,24]. This problem resembles to the well-known 2D Art Gallery problem [25], but in ARM4CH methodology this optimization task will be directed in three dimensions (3D), optimizing the 3D scanner positions. The result is highly dependent on specific constraints set by the user, such as the range of the sensors (scanners), their height related to the ground/floor and their incidence angle. Step 3 terminates with a proposed scanning trajectory/path to be followed by the autonomous and cooperative robotic agents to complete the reality modeling task autonomously. The final path can be created solving a simple Travelling Salesman Problem algorithm, with the identified scanning positions as inputs. The scanning positions near the ground are to be handled by the quadrupedal agents, while those found high above the ground will be tackled by UAVs. The paths/trajectories will be the navigation inputs for cooperative group of robotic agents in the next step.

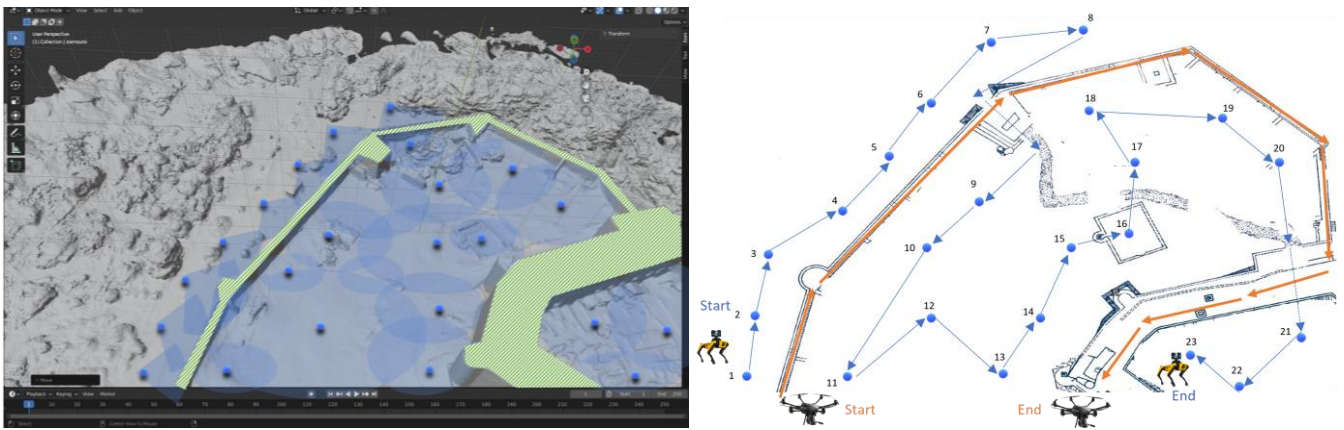

*Figure 4. Calculating the optimum positions (Next Best View) for terrestrial 3D scanning (TLS - blue dots), the uncovered surfaces (UAV – green color) and the respective paths/trajectories (blue and orange arrows).*

It should be also emphasized that if the 3D structure of a monument/site is already available (e.g. Building Information Model BIM) then step 3 may be executed, bypassing steps 1 and 2. In case that there is a prior model of the site available, the next best view simulator may propose the optimal scanning path for the autonomous agents to acquire an updated model of the structure. Figure 5 demonstrates the path for indoor scanning of a complicated monument (Early Christian Orthodox Monastery and Church) based on its respective BIM model.

## 4. *Reality Modeling by unleashing autonomous ground agents (quadrupedals) carrying TLS and UAV in predefined places.*

In this step, the group of robotic agents are programmed to scan the Cultural Heritage site, following the optimum path identified step 3 and performing the minimum required scans in the ground (quadrupedals) and deploying aerial photogrammetry from the air (UAVs). In this stage, the quadrupedal agents carry Terrestrial Laser Scanners (TLS) that provide superior accuracy and precision, while UAVs focus on flight plans specially tailored for the parts of the monument/site that can not be scanned by the TLS in the ground. The result of step 4 is the creation of a more accurate 3D Reality Model of the site/monument surveyed. To this end, step 3 and 4 may be repeatedly executed as many times as needed in order to achieve a coarse-to-fine overall 3D survey, maximizing accuracy, quality and coverage. As a result, the laborious and empirical methodologies that are still implemented based on the operator's expertise and intuition, will be replaced by an automated procedure that is based on quantitative and definite data.

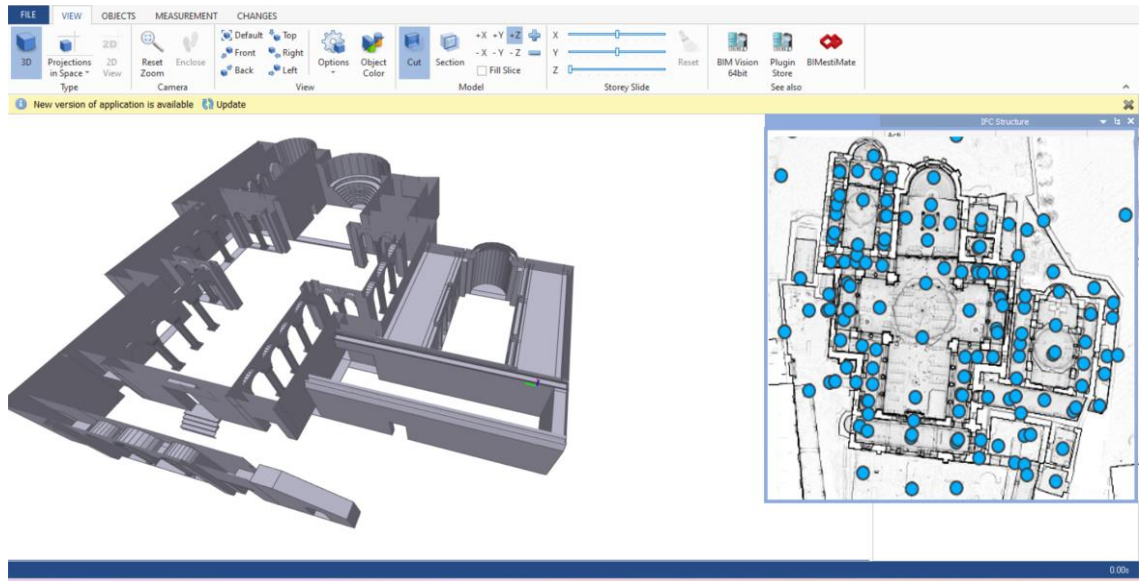

*Figure 5. Solving NBV for TLS for indoor scanning.*

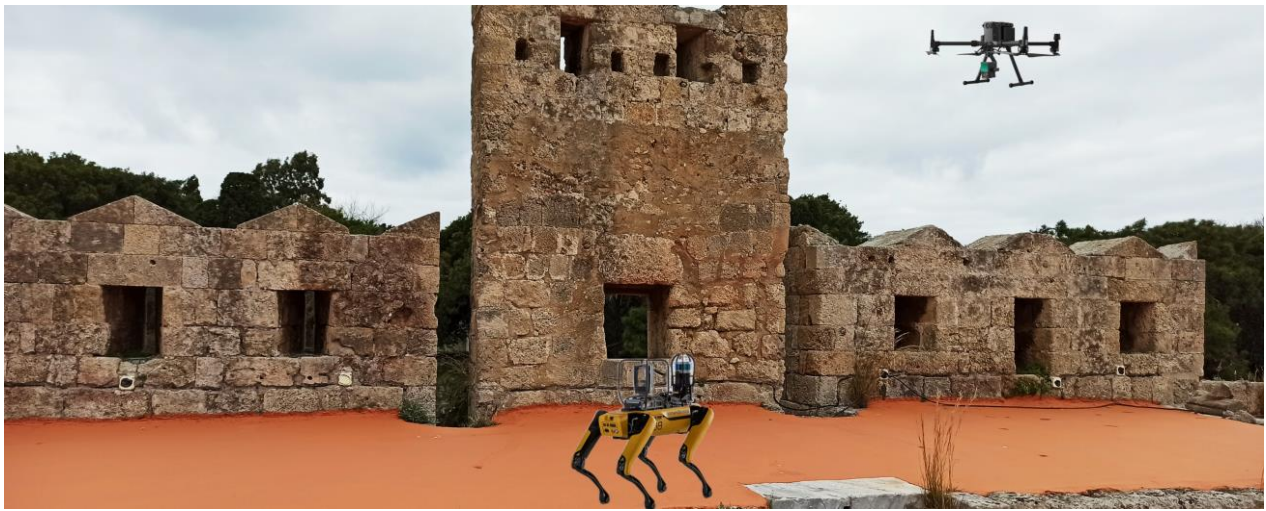

*Figure 6. Revisit and autonomous navigation in CH site, following optimum trajectories identified in step 3 for precise and accurate 3D Reality Modeling.*

## 4. Benefits of the ARM4CH methodology

Using the ARM4CH, researchers/surveyors may send ground/quadrupedal robots on autonomous survey missions (both indoors and outdoors) using SLAM and GPS navigation in full co-operation with aerial vehicles (UAV) for analysis, capture, documentation and 3D scanning. The great benefits exceed the task of Cultural Heritage 3D scanning, since cooperative autonomous Reality Modelling/inspection features the following advantages:

-Schedule robots remotely on unsupervised data capture missions, 24/7, with specific field coverage.

-Ensure accuracy by capturing data from the same locations (viewpoints) multiple times, thus making direct data comparison feasible.

-Ability to create specific schedule plans to capture up-to-date data reliably.

-Reviewing, surveying and inspecting spaces or places of critical/specific importance or those that pose a level of danger to the human surveyor.

-Complement the advantages of various sensor technologies and boosting performance.

-Continuous monitoring, thus once a problem is confirmed, maintenance team may be sent.

From the above it is evident that, a great advantage of ARM4CH methodology is also that it may be replicated/executed systematically, as many times as necessary in forthcoming periods providing the ability of follow-up scans of the same place/site. Those follow-up scans introduce the concept of the fourth dimension (4D) in RM, since now the dimension of time is considered. Consecutive follow-up scans facilitate timeline comparison and monitoring of a constantly changing site and thus flag locations that need emergency actions in times of crisis. To this end ARM4CH may be extremely valuable during the process of establishing and maintaining a Digital Twin (DT) of a CH site or space. This is due to the fact that, " …a DT is a virtual instance of a physical system that is continually updated with the latter's performance" [26] leveraging the most updated available sensor data, to mirror the corresponding physical counterpart.

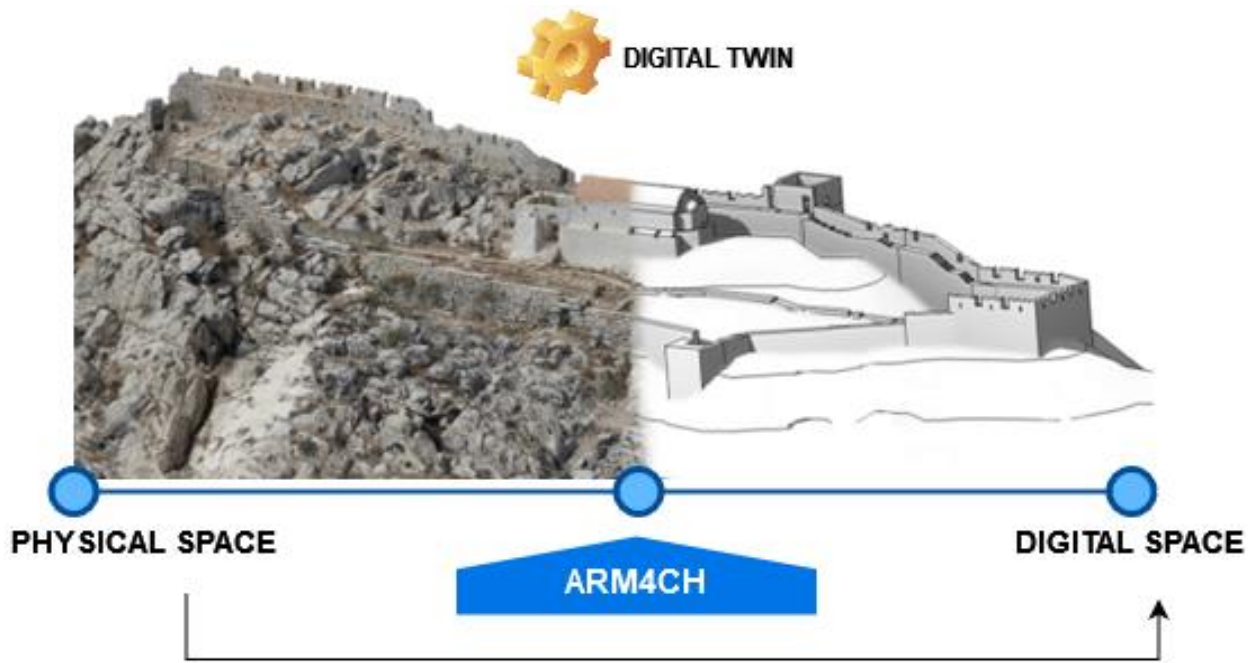

*Figure 7. ARM4CH as a catalyst for continuous model update in the Digital Twin concept.*

## 5. Discussion

In this communication paper, we briefly presented the main steps and stages for a completely new methodology (ARM4CH) to ensure autonomous 3D scanning and digitization for Cultural Heritage spaces. Key enablers of ARM4CH are: a) a technology core platform comprised by autonomous ground quadrupedal robots, as well as UAVs, that work cooperative to navigate and survey large areas using the latest technological sensors, b) the operation of a software visualization tool that resolves the Next Best View problem in 3D meshes and identifies the optimum viewpoint position and scanning path for total survey coverage by ground and UAV robotic agents.

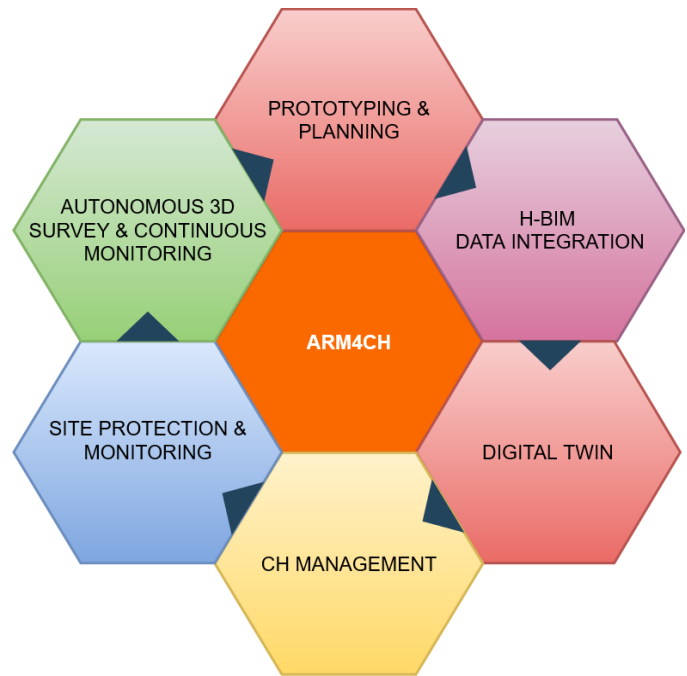

*Figure 8. ARMCH as a core platform for various actions related to Cultural Heritage.*

As already mentioned in Section 4, such a methodology could be essential for a "dynamic" DT of a cultural space to actively respond to the urgent need for efficient management, resilience and sustainability of CH sites, facilities, buildings, structures (indoors and/or outdoors) and their surrounding environment. This undeniable need is emphasized especially in the light of climate change and the necessity of energy saving. The from the unpredictability of hazards must be seen as a duty.

Therefore, since preservation and safeguarding of our Cultural Heritage is an urgent responsibility, there is an increased requirement for automated actions and methodologies to assist preservation, data fusion/integration, site monitoring and holistic management of CH. Using ARM4CH, the "flower" of ARM4CH (see Figure 8) may blossom in critical areas of CH and shift the attention of professionals/experts from a curative towards a more preventive and sustainable approach for CH management.

## 6. References

1. Chen, L., Hoang, D., Lin, H., & Nguyen, T. (2016). Innovative methodology for multi-view point cloud registration in robotic 3d object scanning and reconstruction. Applied Sciences, 6(5), 132. https://doi.org/10.3390/app6050132
2. Sangyoon Park, Sanghyun Yoon, Sungha Ju, Joon Heo, BIM-based scan planning for scanning with a quadruped walking robot, Automation in Construction, Volume 152, 2023, 104911, ISSN 0926-5805, https://doi.org/10.1016/j.autcon.2023.104911
3. P. Kim, J. Park, Y. Cho, As-is geometric data collection and 3D visualization through the collaboration between UAV and UGV, in: Proceedings of the International Symposium on Automation and Robotics in Construction (ISARC) 36, n2019, pp. 544–551, https://doi.org/10.22260/ISARC2019/0073. Banff, Canada, 21–24 May.
4. Peers, C., Motawei, M., Richardson, R., & Zhou, C. (2021). Development of a teleoperative quadrupedal manipulator. UKRAS21 Conference: Robotics at Home Proceedings. https://doi.org/10.31256/hy7sf7g
5. Ding, Y., Pandala, A., Li, C., Shin, Y., & Park, H. W. (2021). Representation-free model predictive control for dynamic motions in quadrupeds. IEEE Transactions on Robotics, 37(4), 1154-1171. https://doi.org/10.1109/tro.2020.3046415

6. Fawcett, R. T., Pandala, A., Ames, A. D., & Hamed, K. A. (2022). Robust stabilization of periodic gaits for quadrupedal locomotion via qp-based virtual constraint controllers. IEEE Control Systems Letters, 6, 1736-1741. https://doi.org/10.1109/lcsys.2021.3133198
7. M. Hutter, C. Gehring, D. Jud, A. Lauber, C.D. Bellicoso,V. Tsounis, J. Hwangbo, K. Bodie, P. Fankhauser, M. Bloesch, R. Diethelm, S. Bachmann, A. Melzer, M. Hoepflinger, "ANYmal - a highly mobile and dynamic quadrupedal robot," 2016 IEEE/RSJ International Conference on Intelligent Robots and Systems (IROS), Daejeon, Korea (South), 2016, pp. 38-44, doi: 10.1109/IROS.2016.7758092.
8. Potthast, C. and Sukhatme, G. (2014). A probabilistic framework for next best view estimation in a cluttered environment. Journal of Visual Communication and Image Representation, 25(1), 148-164. https://doi.org/10.1016/j.jvcir.2013.07.006
9. Bircher, A., Kamel, M., Alexis, K., Oleynikova, H., and Siegwart, R. (2018). Receding horizon path planning for 3D exploration and surface inspection. Auton. Robot., 42(2), 291–306.
10. Delmerico, J. A., Isler, S., Sabzevari, R., & Scaramuzza, D. (2017). A comparison of volumetric information gain metrics for active 3d object reconstruction. Autonomous Robots, 42(2), 197-208. https://doi.org/10.1007/s10514-017-9634-0
11. Almadhoun, R., Abduldayem, A., Taha, T., Seneviratne, L., & Zweiri, Y. (2019). Guided next best view for 3d reconstruction of large complex structures. Remote Sensing, 11(20), 2440. https://doi.org/10.3390/rs11202440
12. Sangyoon Park, Sanghyun Yoon, Sungha Ju, Joon Heo, BIM-based scan planning for scanning with a quadruped walking robot, Automation in Construction, Volume 152, 2023, 104911, ISSN 0926-5805, https://doi.org/10.1016/j.autcon.2023.104911.
13. Hardouin, G., Moras, J., Morbidi, F., Marzat, J., & Mouaddib, E. M. (2020). Next-best-view planning for surface reconstruction of large-scale 3d environments with multiple uavs. 2020 IEEE/RSJ International Conference on Intelligent Robots and Systems (IROS). https://doi.org/10.1109/iros45743.2020.9340897
14. Palazzolo, E. and Stachniss, C. (2018). Effective exploration for mavs based on the expected information gain. Drones, 2(1), 9. https://doi.org/10.3390/drones2010009
15. Kaba, M. D., Uzunbaş, M. G., & Lim, S. (2017). A reinforcement learning approach to the view planning problem. 2017 IEEE Conference on Computer Vision and Pattern Recognition (CVPR). https://doi.org/10.1109/cvpr.2017.541
16. Trummer, M., Munkelt, C., & Denzler, J. (2009). Combined gklt feature tracking and reconstruction for next best view planning. Lecture Notes in Computer Science, 161-170. https://doi.org/10.1007/978-3-642-03798-6_17
17. Wang, Y. and Bue, A. D. (2020). Where to explore next? exhistcnn for history-aware autonomous 3d exploration. Computer Vision – ECCV 2020, 125-140. https://doi.org/10.1007/978-3-030-58526-6_8.
18. Morreale, L., Romanoni, A., & Matteucci, M. (2018). Predicting the next best view for 3d mesh refinement. Intelligent Autonomous Systems 15, 760-772. https://doi.org/10.1007/978-3-030-01370-7_59
19. M. Kazhdan, M. Bolitho, H. Hoppe, Poisson Surface Reconstruction, Proceedings of the fourth Eurographics symposium on Geometry, 2006 61-70, 2006.
20. S. Kim, H. G. Kim, T. Kim, Mesh modelling of 3d point cloud from UAV images by point classification and geometric constraints, Proc. of International Archives of the Photogrammetry, Remote Sensing and Spatial Information Sciences, Volume XLII-2, 2018 ISPRS TC II Mid-term Symposium "Towards Photogrammetry 2020", pp. 507-511, https://doi.org/10.5194/isprs-archives-XLII-2-507-2018.
21. Liao, Yiyi, Simon Donne, and Andreas Geiger. "Deep marching cubes: Learning explicit surface representations." Proceedings of the IEEE Conference on Computer Vision and Pattern Recognition. 2018.
22. Lorensen, William E., and Harvey E. Cline. "Marching cubes: A high resolution 3D surface construction algorithm." Seminal graphics: pioneering efforts that shaped the field. 1998. 347-353.
23. Robin, C., Lacroix, S., Multi-robot target detection and tracking: taxonomy and survey. Auton. Robots 40 (4), 2016, 729–760.

24. Li, Y.F., Liu, Z.G., Information entropy-based viewpoint planning for 3-D object reconstruction. IEEE Trans. Robot. 21 (3), 2015, 324–337.
25. O'Rourke, J., 1987. Art Gallery Theorems and Algorithms. Oxford University Press.
26. A. Madni, C. Madni, and S. Lucero, ''Leveraging digital twin technology in model-based systems engineering,'' Systems, vol. 7, no. 1, p. 7, Jan. 2019.